\documentclass{ISMA_USD}
\usepackage{subcaption}
\usepackage{placeins}
\usepackage[numbers,sort&compress]{natbib}

\hypersetup{pdftitle  = {When does a bridge become an aeroplane?},
	pdfauthor = {{T.A.\ Dardeno}, {L.A.\ Bull}, N. Dervilis, K. Worden},
	pdfkeywords = {ISMA2024, USD2024, template}}

\title{When does a bridge become an aeroplane?}

\author[1]{{T.A.\ Dardeno}}
\author[2,3]{{L.A.\ Bull}}
\author[1]{N. Dervilis}
\author[1]{K. Worden}
\affil[1]	{Dynamics Research Group, Department of Mechanical Engineering, University of Sheffield, \NewLineAffil Mappin Street, Sheffield S1 3JD, UK \NewLineAffil e-mail: \textbf{t.a.dardeno@sheffield.ac.uk} \NewLineAffil  } 

\affil[2]	{Department of Engineering, University of Cambridge, Cambridge CB3 0FA, UK \NewLineAffil }

\affil[3]	{School of Mathematics and Statistics, University of Glasgow, Glasgow G12 8SQ, Scotland}
		
\date{}

\begin{document}
%\bstctlcite{IEEEtran:BSTadapt}	

%%%%%%%%%%%%
% Abstract %
%%%%%%%%%%%%

% Enter your abstract in the command below.

\abstract{Despite recent advances in population-based structural health monitoring (PBSHM), knowledge transfer between highly-disparate structures (i.e., heterogeneous populations) remains a challenge. It has been proposed that heterogeneous transfer may be accomplished via intermediate structures that bridge the gap in information between the structures of interest. A key aspect of the technique is the idea that by varying parameters such as material properties and geometry, one structure can be continuously morphed into another. The current work demonstrates the development of these interpolating structures, via case studies involving the parameterisation of (and transfer between) a simple, simulated `bridge' and `aeroplane'. The facetious question `When is a bridge not an aeroplane?' has been previously asked in the context of predicting positive transfer based on structural similarity. While the obvious answer to this question is `Always,' the current work demonstrates that in some cases positive transfer can be achieved between highly-disparate systems.}

%	While many recent advances in population-based structural health monitoring (PBSHM) have been made to improve information transfer between similar structures (i.e., homogeneous populations), transfer between structures with highly-disparate features (i.e., heterogeneous populations) remains a challenge. It has been proposed that transfer within a heterogeneous population may be accomplished via intermediate structures that bridge the gap in information between the structures of interest. A key aspect of the technique is the idea that by varying parameters such as material properties, geometry, and boundary stiffness, one structure can be continuously morphed into another. The current work demonstrates the development of these interpolating structures, via case studies involving the parameterisation of (and transfer between) a simple, simulated `bridge' and `aeroplane'. The facetious question `When is a bridge not an aeroplane?' has been previously asked in the context of predicting positive transfer based on structural similarity. While the obvious answer to this question is `Always,' the current work demonstrates that in some cases positive transfer can be achieved between highly-disparate systems.

\maketitle

%%%%%%%%%%%%%%%%%%%%%%%%%%
% Beginning of the paper %
%%%%%%%%%%%%%%%%%%%%%%%%%%

\section{Introduction}

An exciting prospect for addressing the challenge of transfer between heterogeneous structures involves leveraging the inherent geometry underlying the space of structures. Traditional linear machine-learning methods typically struggle with non-Euclidean data \cite{bronstein2017geometric, nonEuclideanML}, whereas geometric approaches \cite{Tsialiamanis2021,gopalan2011domain,Boqing2012,masci2015geodesic,monti2017geometric,asif2021graph,simon2021learning} are well-suited for navigating the intricate, curved manifold structures of non-Euclidean spaces. In addition, in areas outside of SHM, implementing intermediate steps in the transfer process has been shown to facilitate smoother transitions between vastly different domains or tasks \cite{gopalan2011domain,Boqing2012,rusu2022progressive,sagawa2022gradual,simon2021learning}.

To clarify; one of the ideas in population-based structural health monitoring (PBSHM) is that the structures of a given population can be expressed abstractly in the form of an attributed graph, which allows them to be embedded in a metric space -- a space of graphs \cite{Tsialiamanis2021}. For two structures $S$ and $S'$, given the metric space structure, one can calculate the {\em distance} $d(S,S')$ between them. If the calculated distance were to be lower than some threshold $\epsilon_d$, the PBSHM framework would dictate that transfer may be attempted. 

An important issue, then, is how transfer can be achieved when the distance between two structures of interest is too large for positive transfer. Suppose the task is to transfer to a new structure $S$, which is data-poor, but there is no structure $S'$ in the current population for which $d(S,S') \le \epsilon_d$. Recall that PBSHM does not distinguish (in its representation space), between real structures and models; as such, a model {\em intermediate} structure $S^*$ may be constructed for which $d(S,S*) \le \epsilon_d$ and $d(S',S*) \le \epsilon_d$. In this situation, transfer may be accomplished in two steps; first from $S'$ to $S^*$ and then from $S^*$ to $S$. Furthermore, for large distances between $S$ and $S'$, multiple intermediate structures may be developed, to enable transfer via a greater number of steps. It is important to note that while transfer is carried out in the feature spaces of the structures, it can be argued that proximity in the structure space is equivalent to proximity in the data space \cite{Tsialiamanis2021}. In transfer-learning terms, the feature spaces of $S$ and $S'$ are the {\em target} and {\em source} domains, respectively. 

Transfer can be considered to be a map between data domains. Geodesic flows, \cite{gopalan2011domain,Boqing2012}, which are derived from differential geometry, identify the shortest path between two domains by leveraging the underlying geometry of the space. Gopalan \emph{et al.} \cite{gopalan2011domain} used a geodesic-flows approach in the context of unsupervised domain adaptation for object recognition, representing the source and target domains as subspaces on a Grassmannian manifold. The approach in \cite{gopalan2011domain} is influenced by incremental learning \cite{VanderVanIncLearn,KashefIncLearn}, and involves identifying potential intermediate domains between the source and target and using a finite number of these domains to learn domain transitions. Building upon the work in \cite{gopalan2011domain}, Gong \emph{et al.} \cite{Boqing2012} later introduced the geodesic flow kernel, which integrates an infinite series of subspaces along the flow, for improved domain-shift modelling.

In accordance with these principles, a \emph{heterogeneous} transfer approach for PBSHM was presented in \cite{DardenoEWSHM}. Given that proximity in the structure space typically corresponds to that in the data space, a continuous chain of structures was generated between simple, simulated one- and two-support bridges, via parameterisation of the material properties and position of the second support. Direct transfer of damage labels from the one-support bridge to the two-support bridge was poor; however, excellent results were achieved by transferring across several intermediate structures. The current work considers the more ambitious task of transferring between a simulated `bridge' and `aeroplane', via a chain of intermediate structures generated by parameterising the geometry, material properties, and boundary stiffness of the models. Transfer learning along the chain is performed via normal-condition alignment \cite{PooleNCA} with classification using support vector machines (SVM) \cite{SVM1992} first with a linear kernel, and then with the geodesic flow kernel \cite{Boqing2012}. The facetious question `When is a bridge not an aeroplane?' has been asked with respect to predicting positive transfer based on structural similarity \cite{worden2021bridge}. While the obvious answer to the question is `Always,' the current work demonstrates that in some cases positive transfer can be achieved between highly-disparate structures.

The layout of this paper is as follows. Section 2 provides an overview of the theoretical background of the proposed work, including the geodesic flow kernel. Section 3 discusses the case studies using simulated structures, with conclusions in Section 4. 

\section{Theoretical Background}

Included in this section is a brief introduction to the geodesic flow kernel and the associated equations required for its formulation; for a more comprehensive discussion of geodesic flows and the geodesic flow kernel, interested readers are directed to \cite{gopalan2011domain,Boqing2012}.
	
\subsection{Geodesic flow kernel}

The geodesic flow kernel (GFK) \cite{Boqing2012}, characterises incremental changes in geometrical and statistical properties between the source and target domains via integration of all subspaces along the flow. To construct the kernel, PCA subspaces are computed and their appropriate dimensionality determined. The principal angles of the subspaces are then used to develop the geodesic flow. The geodesic flow kernel is then constructed and embedded into a kernel-based classifier \cite{Boqing2012}. 

Let $ \mathbb{G}( \text{D}, \text{d} ) $ represent the Grassmannian manifold, which is the collection of all $d$-dimensional subspaces of $\mathbb{R}^{\text{D}}$. Let $\boldsymbol{S}_1,\boldsymbol{S}_2 \in \mathbb{R}^{\text{D} \times \text{d}} $ signify the principal component analysis (PCA) \cite{PCA1987} bases of the source and target, respectively. Then, let $\boldsymbol{R}_1 \in \mathbb{R}^{\text{D} \times (\text{D}-\text{d})}$ and  $\boldsymbol{Q} \in \mathbb{R}^{\text{D} \times \text{D}} $ define the orthogonal complement and orthogonal completion of $\boldsymbol{S}_1$, respectively. The cosine-sine decomposition of $\boldsymbol{Q} ^{\intercal} \boldsymbol{S}_2$ is given by,

\begin{equation}
	\boldsymbol{Q}^{\intercal} \boldsymbol{S}_2 = \begin{bmatrix}
		\boldsymbol{V}_1 & 0 \\
		0 & \boldsymbol{\tilde{V}}_2 
	\end{bmatrix}
	\begin{bmatrix}
		\bold{\Gamma} \\
		-\bold{\Sigma} 
	\end{bmatrix}
	\boldsymbol{V}^{\intercal}
\end{equation}

\noindent 
where $\boldsymbol{V}_1$, $\boldsymbol{\tilde{V}}_2$, and $\boldsymbol{V}$ are orthogonal matrices that rotate/align the subspaces onto a common basis, such that $\boldsymbol{S}_1^{\intercal} \boldsymbol{S}_2 = \boldsymbol{V}_1 \bold{\Gamma} \boldsymbol{V}^{\intercal}$ and $\boldsymbol{R}_1^{\intercal} \boldsymbol{S}_2 = -\boldsymbol{V}_2 \bold{\Sigma} \boldsymbol{V}^{\intercal}$ \cite{gopalan2011domain,Boqing2012}. The arccosine and arcsine of matrices $\bold{\Gamma}$ and $\bold{\Sigma}$ are used to compute the principal angles, ${\theta}$, respectively \cite{gopalan2011domain,Boqing2012}, which are then used to develop the geodesic flow. Via the canonical Euclidean metric on the Riemannian manifold, the geodesic flow is parameterised as $ \bold{\Phi} : t \in [0,1] \rightarrow \bold{\Phi}(t) \in \mathbb{G}(\text{d},\text{D})$, with the constraints that $ \bold{\Phi}(0) = \boldsymbol{S}_1$ and  $\bold{\Phi}(1) = \boldsymbol{S}_2$ \cite{gopalan2011domain,Boqing2012}. For other $t$, $\bold{\Phi}(t)$ can be given as \cite{gopalan2011domain,Boqing2012},

\begin{equation}
	\bold{\Phi}(t) = \boldsymbol{Q} \begin{bmatrix}
		\boldsymbol{V}_1 \bold{\Gamma}(t) \\
		-\boldsymbol{\tilde{V}}_2\bold{\Sigma}(t)
	\end{bmatrix}
\end{equation}

Now, assume two $D$-dimensional feature vectors $\boldsymbol{x}_i$ and $\boldsymbol{x}_j$, whose projections into the space defined by $\bold{\Phi}(t)$ are calculated for continuous time $t$ from 0 to 1 \cite{Boqing2012}. These projections are then concatenated to form the infinite-dimensional feature vectors $\boldsymbol{z}_i^\infty$ and $\boldsymbol{z}_j^\infty$ \cite{Boqing2012}. Via the kernel trick, the inner product between these vectors gives the geodesic flow kernel, $\boldsymbol{G}$,

\begin{equation}
	\langle \boldsymbol{z}_i^\infty, \boldsymbol{z}_j^\infty \rangle = \int_{0}^{1} \left( \boldsymbol{\Phi}(t)^T \boldsymbol{x}_i \right)^T \left( \boldsymbol{\Phi}(t)^T \boldsymbol{x}_j \right) dt = \boldsymbol{x}_i^T \boldsymbol{G} \boldsymbol{x}_j
\end{equation}

\noindent 
where $\boldsymbol{G} \in \mathbb{R}^{\text{D} \times \text{D}} $ is a positive semidefinite matrix \cite{Boqing2012}. The matrix $\boldsymbol{G}$ can be written in closed form \cite{Boqing2012},

\begin{equation}
	\boldsymbol{G} = \boldsymbol{Q} 
	\begin{bmatrix}
		\boldsymbol{V}_1 & 0 \\
		0 & -\boldsymbol{\tilde{V}}_2
	\end{bmatrix}
	\begin{bmatrix}
		\bold{\Lambda}_1 & \bold{\Lambda}_2 \\
		\bold{\Lambda}_2 & \bold{\Lambda}_3 
	\end{bmatrix}
	\begin{bmatrix}
		\boldsymbol{V}_1^{\intercal} & 0 \\
		0 & -\boldsymbol{\tilde{V}}_2^{\intercal} 
	\end{bmatrix}
	\boldsymbol{Q}^{\intercal}
\end{equation}

\noindent 
where $\bold{\Lambda}_1$, $\bold{\Lambda}_2$, and $\bold{\Lambda}_3$ are diagonal matrices with elements \cite{Boqing2012},

\begin{equation}
	\lambda_{1i} = 1 + \frac{\sin(2\theta_i)}{2\theta_i}, \quad \lambda_{2i} = \frac{\cos(2\theta_i) - 1}{2\theta_i}, \quad \lambda_{3i} = 1 - \frac{\sin(2\theta_i)}{2\theta_i}
\end{equation}

\begin{figure}[h!]
	\vspace{0.5cm}
	\centering
	\includegraphics[width=0.95\textwidth]{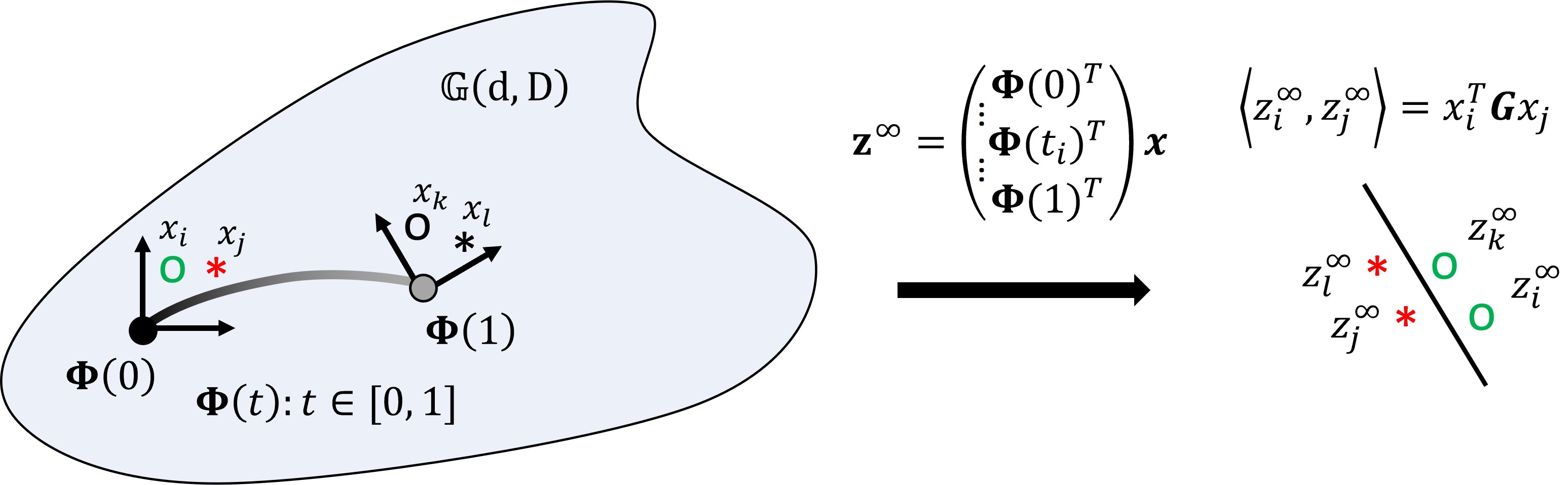}
	\caption{Geodesic flow kernel, adapted from \cite{Boqing2012}.}
	\label{fig:GFK}
	% \vspace{-0.5cm}
\end{figure}

\noindent This process is shown in Figure \ref{fig:GFK}. It is important to note that the geodesic flow kernel measures similarity by considering information from both the source and target domains. As such, the GFK is insensitive to smooth domain shifts, and can therefore provide better transfer compared to traditional methods. The GFK is used in this work to facilitate transfer between two highly-disparate structures by embedding it into a support vector machine (SVM) classifier, as discussed in Section 3.

\section{From a bridge to an aeroplane}
The intent of this work is to demonstrate how in some cases, it may be possible to treat highly-disparate structures as differing only in their values for a certain set of parameters. Models of these structures can be generated, and varying these parameters within a given interval can result in a continuous and gradual morphing of one structure into another. A subset of these models can be used to facilitate information transfer, by incrementally transferring along the chain. This section discusses the development of parametric models that incrementally span between simple bridge and aeroplane structures (Section 3.1), the incremental transfer approach used (Section 3.2), and the results of the associated case studies (Section 3.3). Note that these case studies use only simulated structures and data.

\subsection{Model development}
Finite element models were generated using PyAnsys. The source structure, $ \boldsymbol{S}_1 $, was a four-span bridge, with a concrete 100-metre-long, 15-metre-wide, and 2-metre-thick deck comprised of concrete. Properties of masonry were used for the three 15-metre-tall, 2-metre-wide, and 2-metre-thick supports. Ground springs ($10\times10^{10}$ N/m) were used to simulate a fixed boundary condition at both ends of the deck. At the base of the supports, ground springs ($10\times10^{10}$ N/m) were used to restrict motion in the forward and lateral directions and a fixed boundary condition was applied in the vertical direction (the fixed support was applied instead of springs as it would not change across the structures). The target structure, $ \boldsymbol{S}_2 $,  was a highly-simplified aluminium `aeroplane' based on the benchmark by the Structures and Materials Action Group (SM-AG19) of the Group for
Aeronautical Research Technology in EURope (GARTEUR). The structure was modelled with an 18-metre wingspan, where each wing was 3-metres wide and 0.5-metres thick. The 20-metres-long fuselage was 2-metres wide and 2-metres in height. Each wing and the fuselage stood on a 4-metre-tall, 0.5-metre-wide, and 0.5-metre-thick `landing gear' support. The wings were allowed to move freely (ground spring stiffness 0 N/m), and the base of the supports were constrained only in the vertical direction. The material properties, dimensions, and ground spring stiffness were then varied between those for $ \boldsymbol{S}_1 $, and those for $ \boldsymbol{S}_2 $, to generate a continuous series of intermediate structures between them. A total of 80 models were generated, and all were meshed using 10-noded tetrahedral elements. Material properties and dimensions are listed in Table \ref{tab:Properties}. Note that these values are used here for demonstration purposes but can be modified and refined as needed. 

\begin{figure}[ht!]
	\centering
	\begin{subfigure}{.5\textwidth}
		\centering
		\includegraphics[width=\linewidth, trim={16cm 12.5cm 16cm 12.5cm}, clip]{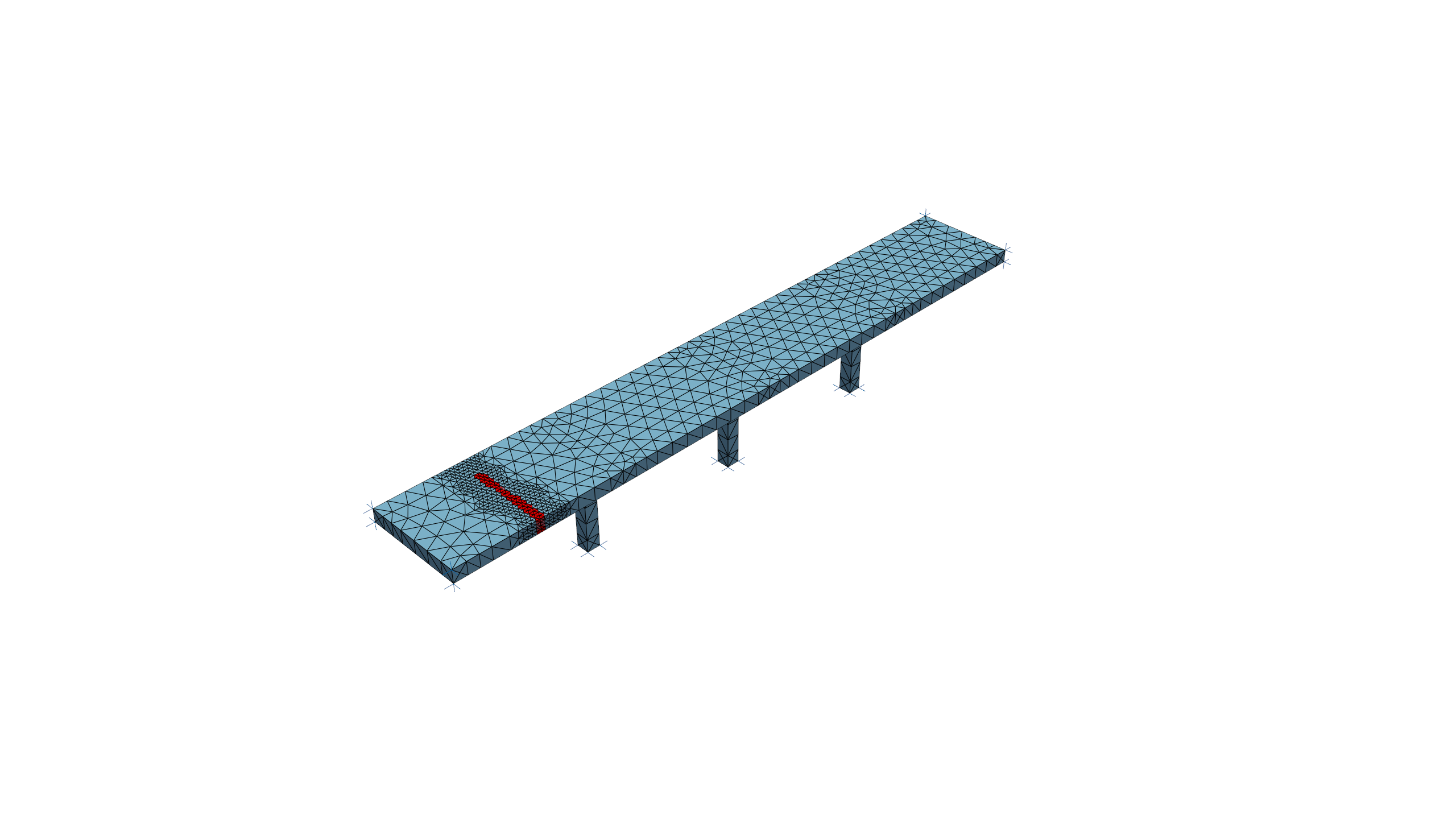}
	\end{subfigure}%
	\begin{subfigure}{.49\textwidth}
		\centering
		\includegraphics[width=\linewidth, trim={16cm 12.5cm 16cm 12.5cm}, clip]{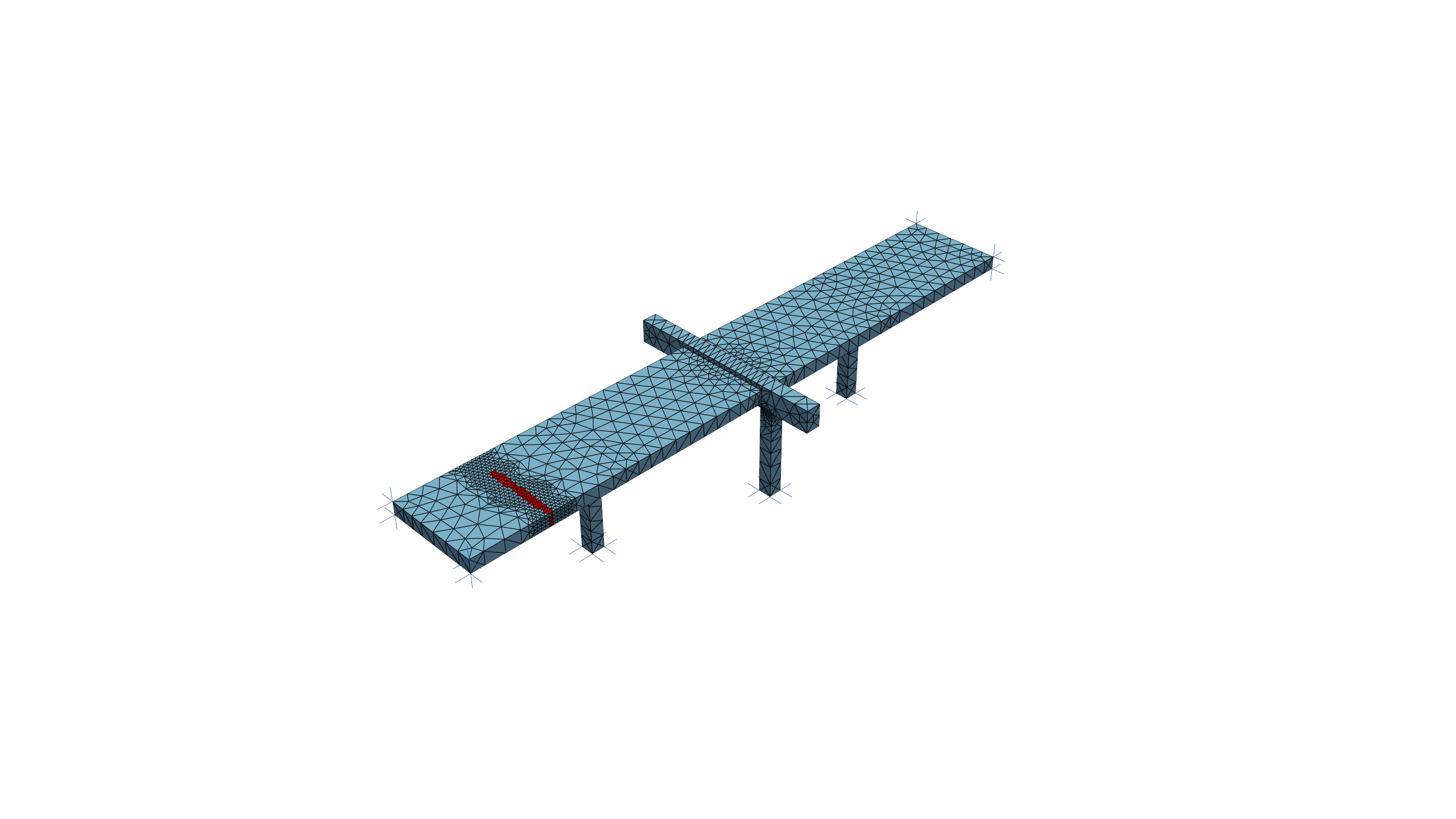}
	\end{subfigure} \\
	\begin{subfigure}{.49\textwidth}
		\centering
		\includegraphics[width=\linewidth, trim={16cm 12.5cm 16cm 12.5cm}, clip]{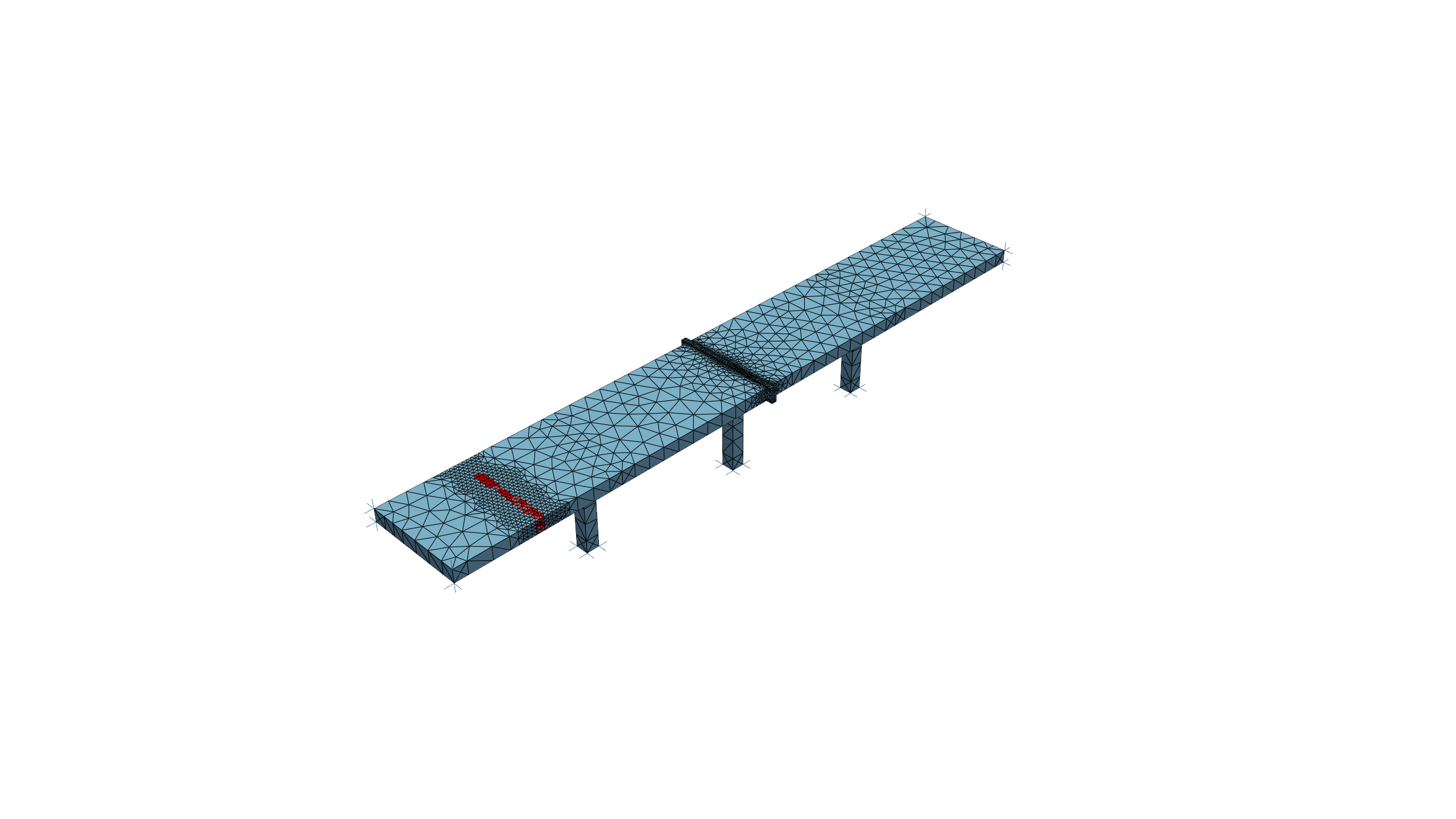}
	\end{subfigure} 
	\begin{subfigure}{.49\textwidth}
		\centering
		\includegraphics[width=\linewidth, trim={16cm 12.5cm 16cm 12.5cm}, clip]{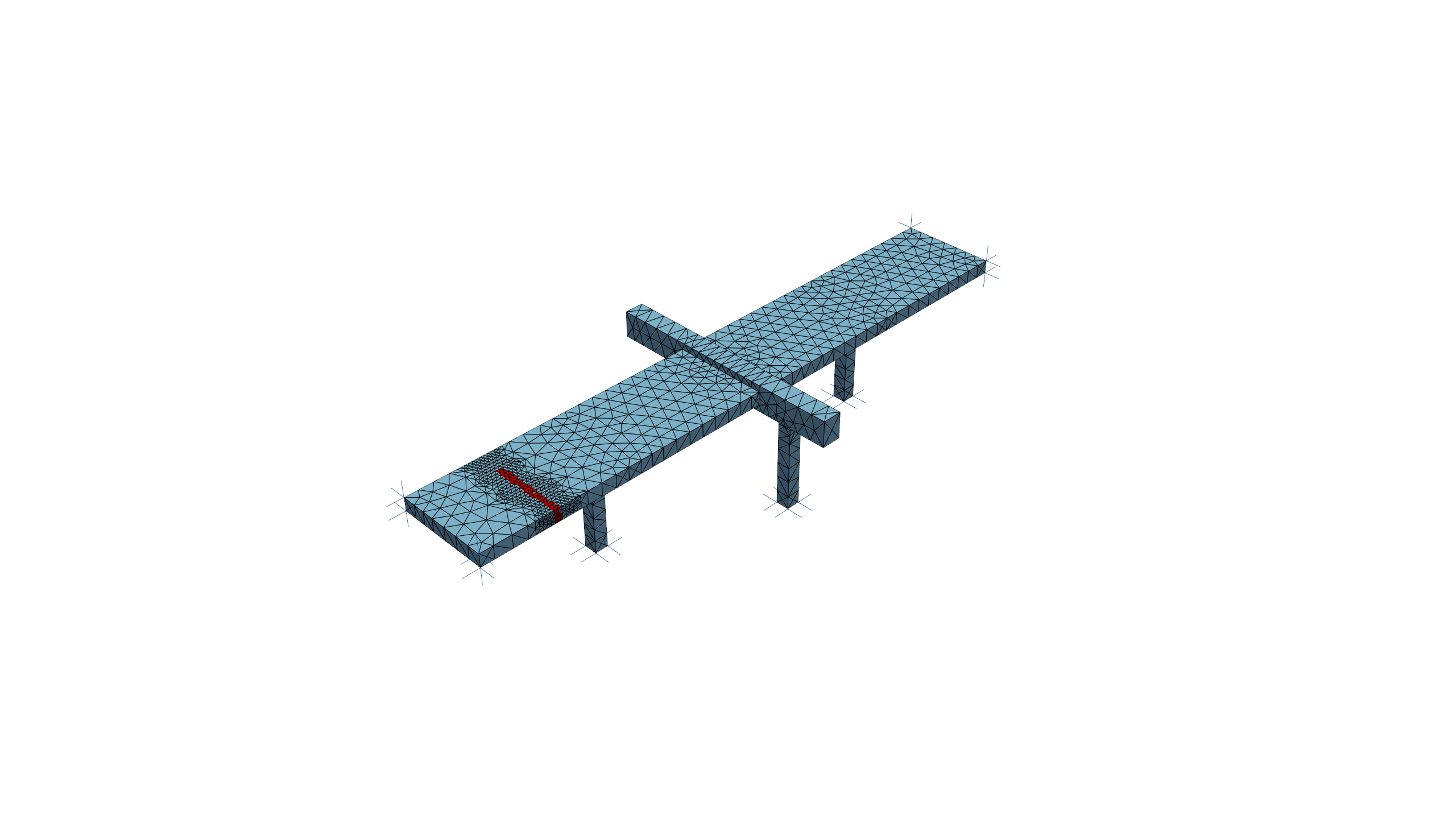}
	\end{subfigure} \\
	\begin{subfigure}{.49\textwidth}
		\centering
		\includegraphics[width=\linewidth, trim={16cm 12.5cm 16cm 12.5cm}, clip]{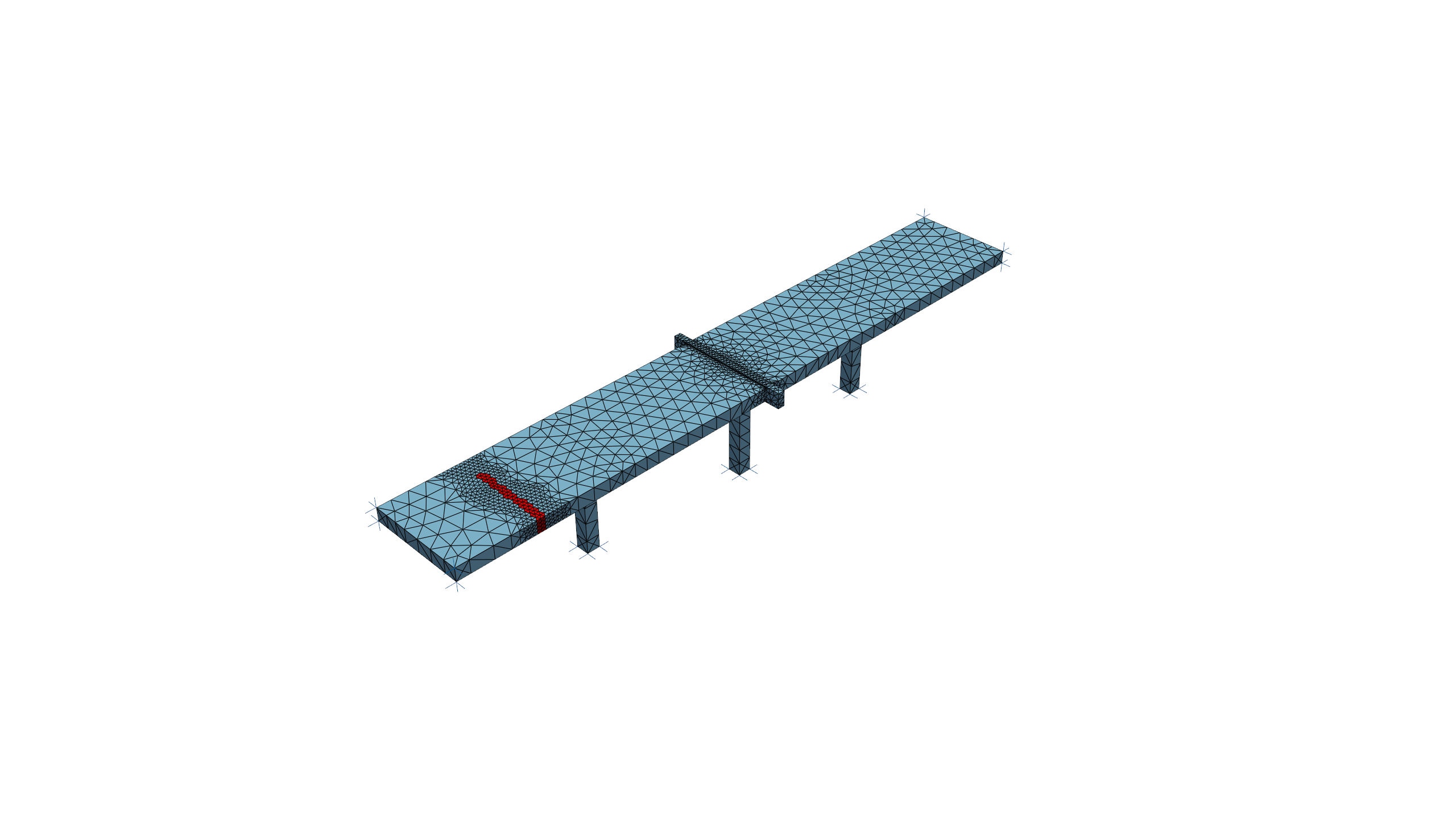}
	\end{subfigure}
	\begin{subfigure}{.49\textwidth}
		\centering
		\includegraphics[width=\linewidth, trim={16cm 12.5cm 16cm 12.5cm}, clip]{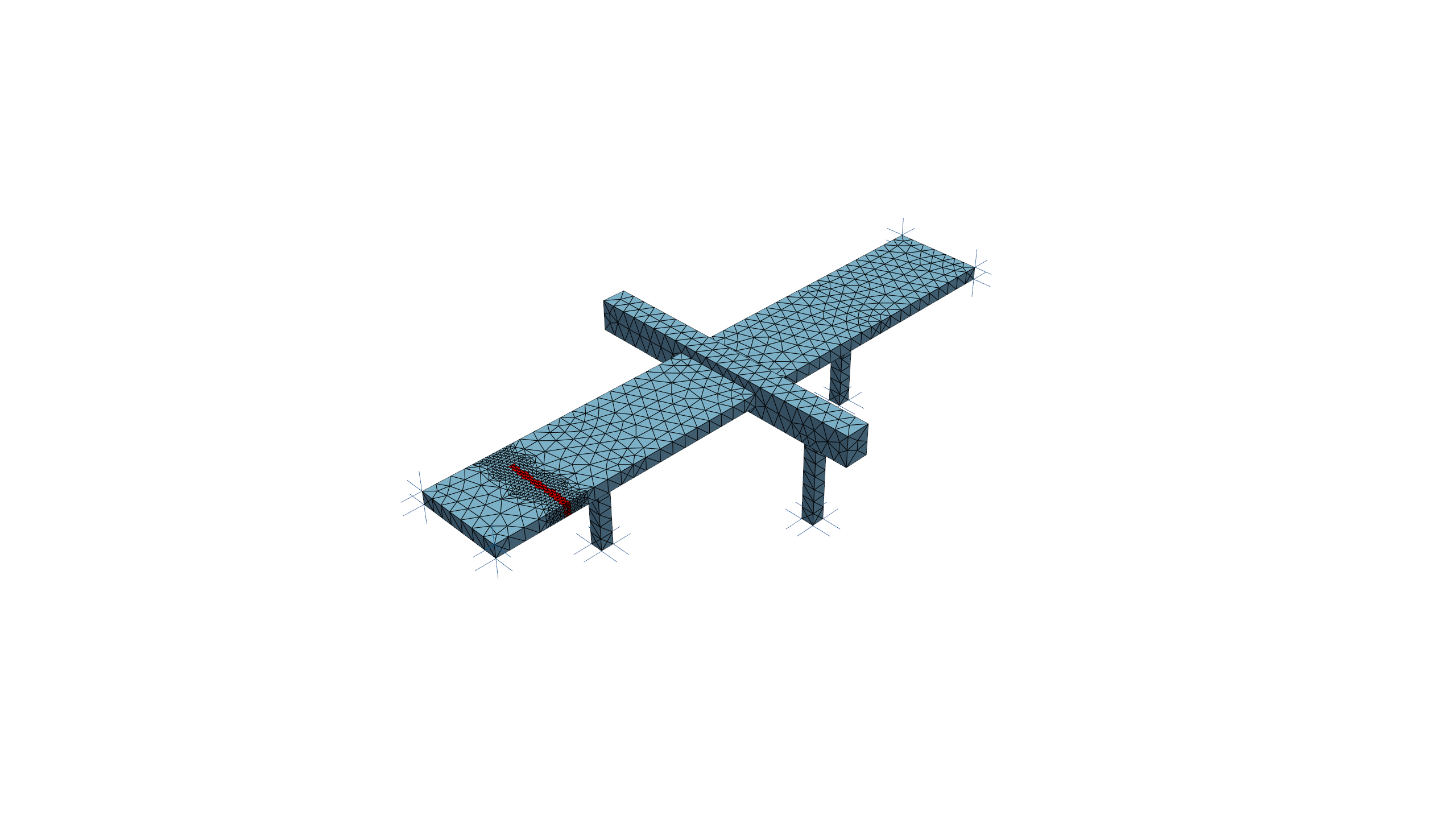}
	\end{subfigure}
	\begin{subfigure}{.49\textwidth}
		\centering
		\includegraphics[width=\linewidth, trim={16cm 12.5cm 16cm 12.5cm}, clip]{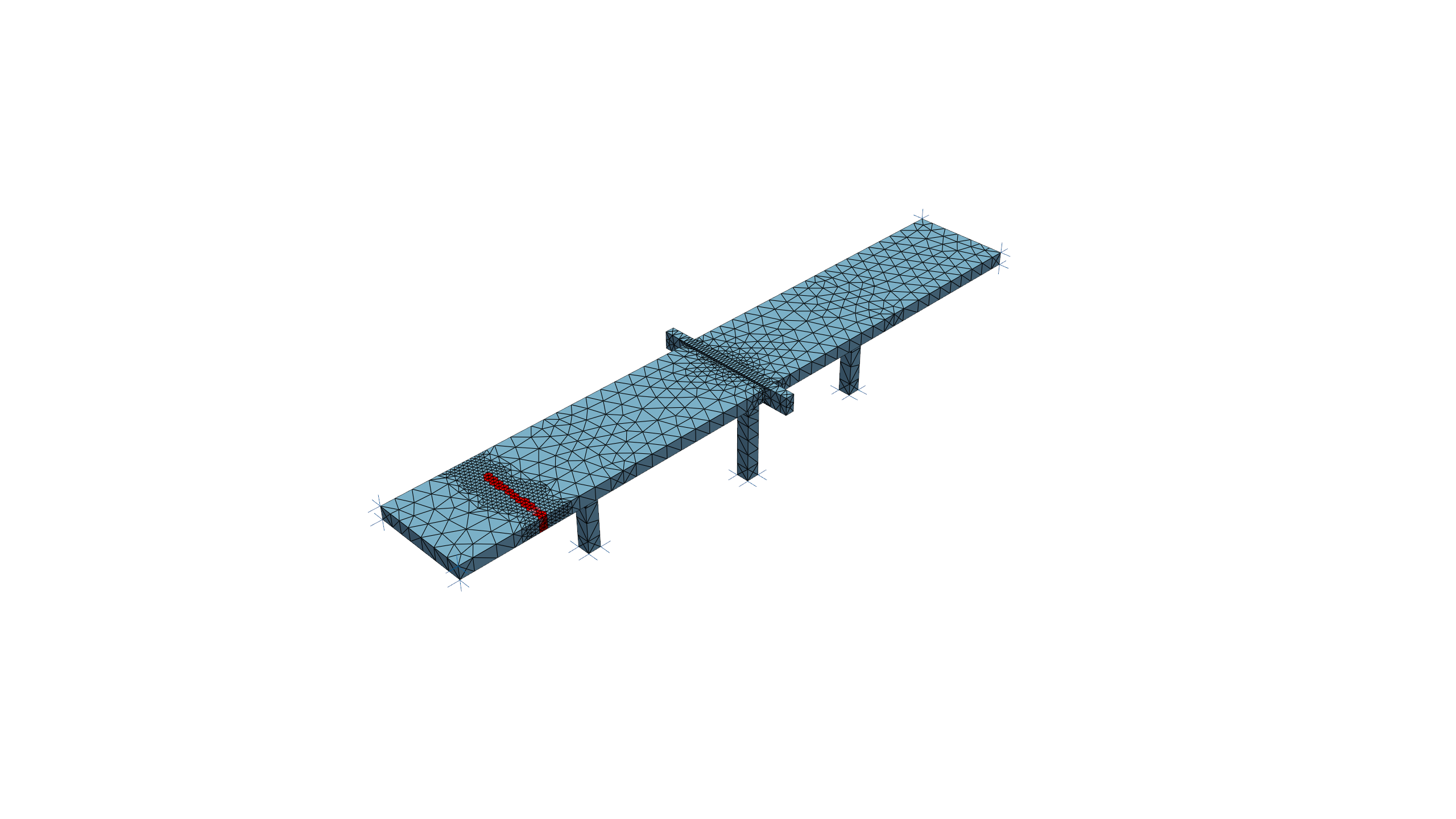}
	\end{subfigure}
	\begin{subfigure}{.49\textwidth}
		\centering
		\includegraphics[width=\linewidth, trim={16cm 12.5cm 16cm 12.5cm}, clip]{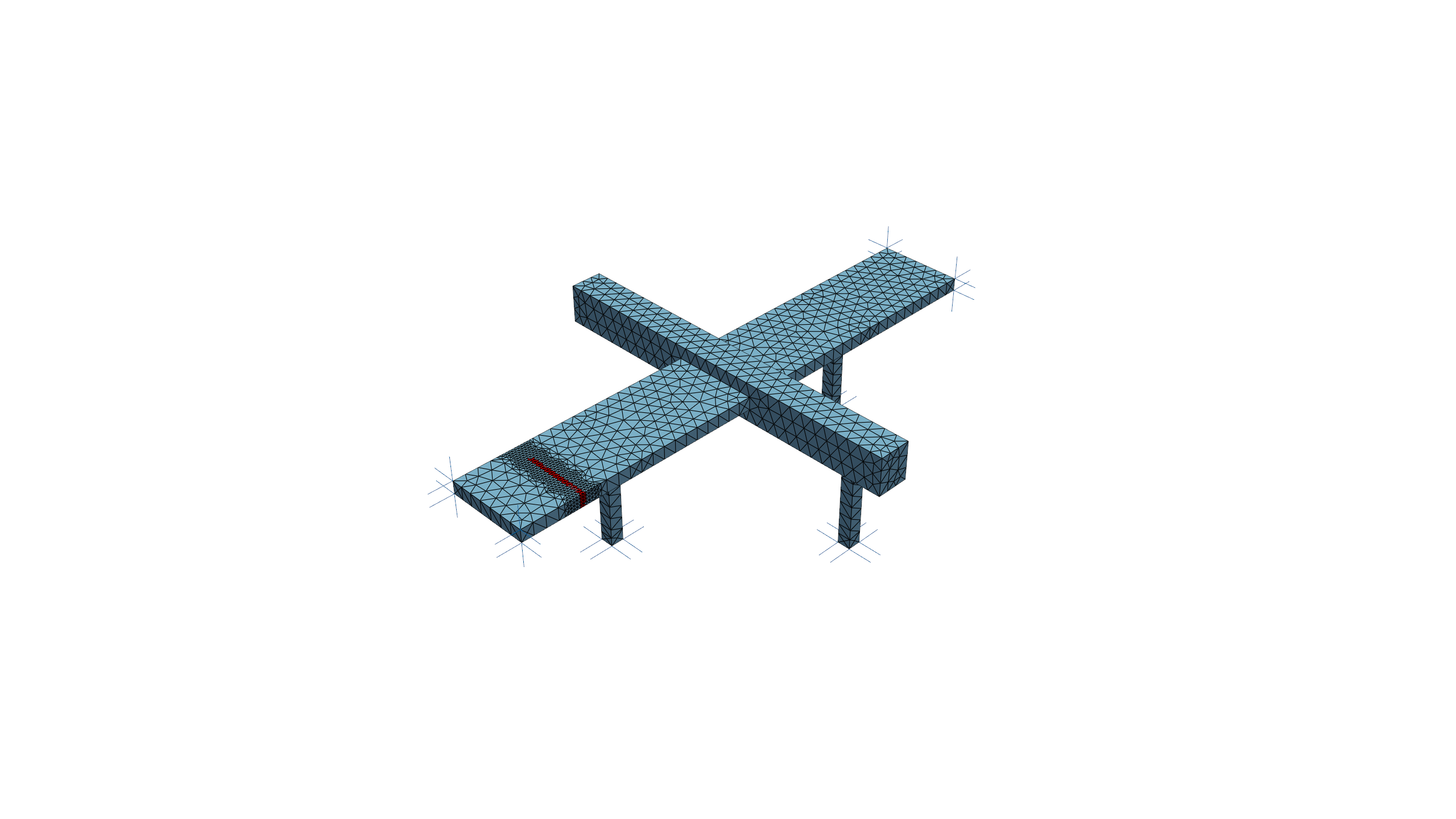}
	\end{subfigure}
	\begin{subfigure}{.49\textwidth}
		\centering
		\includegraphics[width=\linewidth, trim={16cm 12.5cm 16cm 12.5cm}, clip]{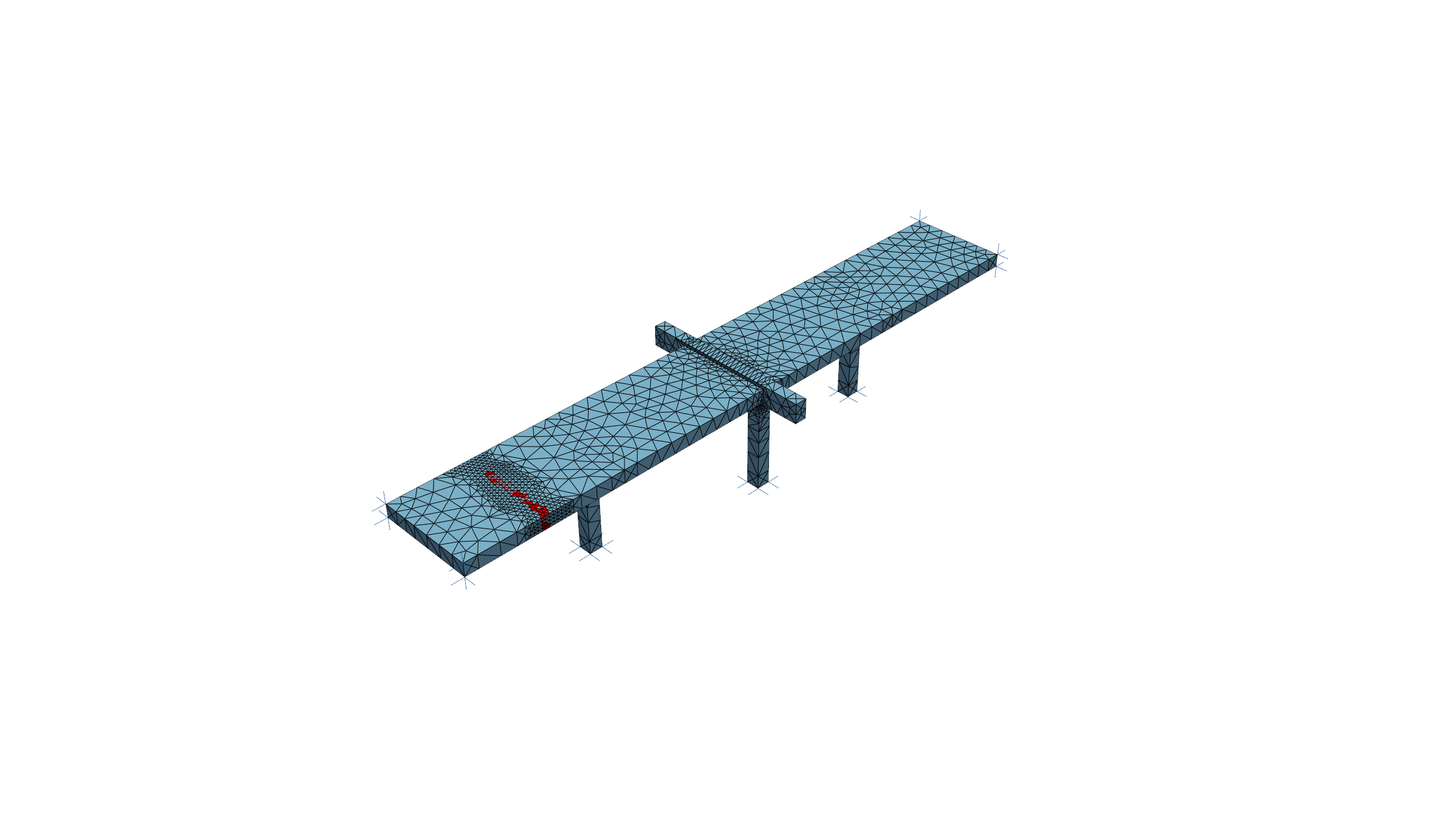}
	\end{subfigure}
	\begin{subfigure}{.49\textwidth}
		\centering
		\includegraphics[width=\linewidth, trim={16cm 12.5cm 16cm 12.5cm}, clip]{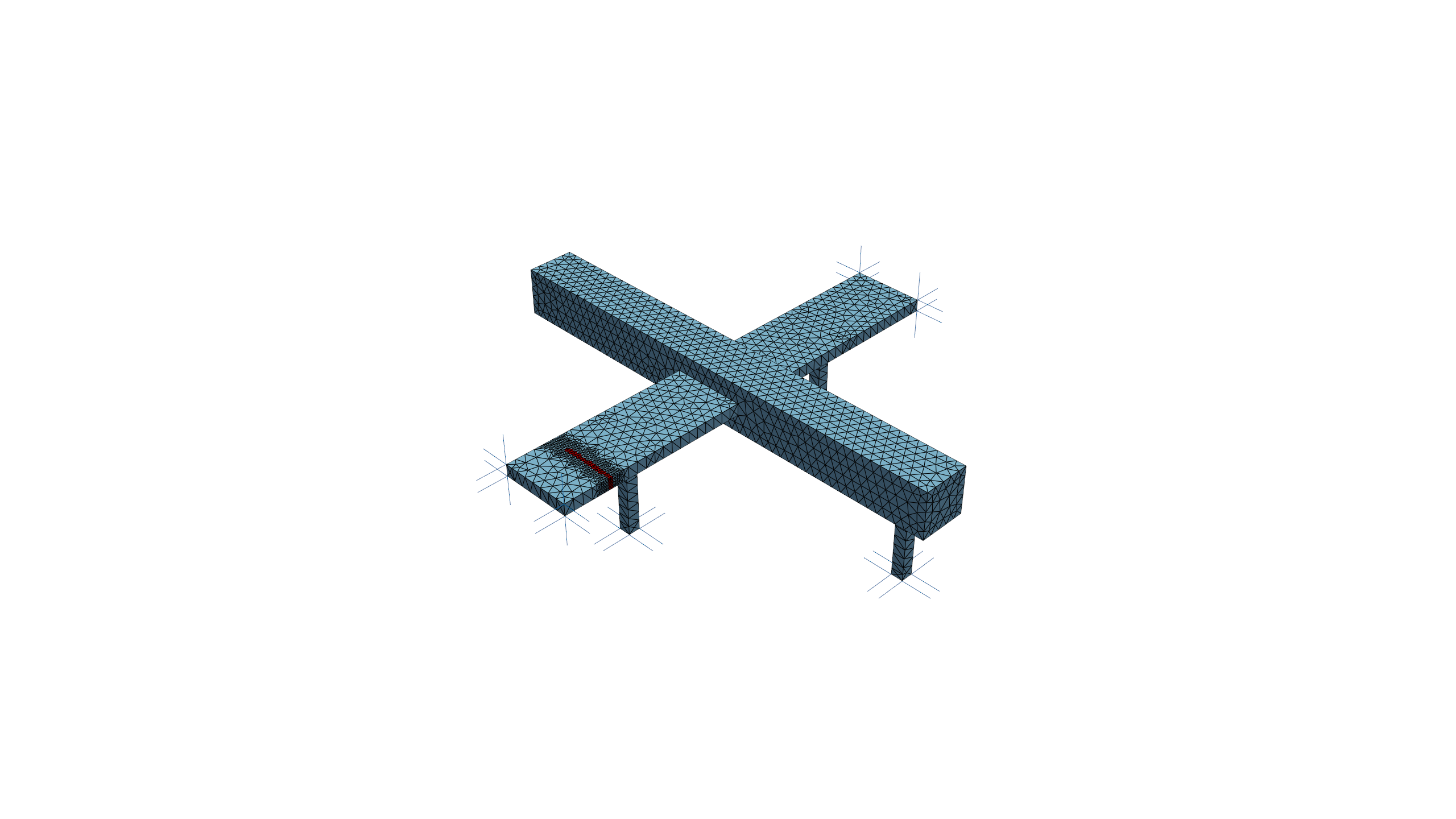}
	\end{subfigure}
	\caption{Model generation with intermediate states and simulated crack (in red).}
	\label{fig:models}
\end{figure}

For all structures, normal-condition datasets were generated using the first 15 natural frequencies and replicating them each 100 times with added noise proportional to the frequency. To simulate a crack, the Young's Modulus of a small section of elements located on the far left span/wing was reduced. The relative size and location of the crack was consistent throughout the transformations. Damage-condition datasets were then generated in the same manner as the healthy datasets. A subset of the intermediate models (with the crack shown in red) is presented in Figure \ref{fig:models}.

\begin{table}[h]
	\renewcommand{\arraystretch}{1.5}
	\caption{Model Properties.}
	\label{tab:Properties}
	\centering 
	\begin{tabular}{ccccccc}
		\hline
		& \multicolumn{2}{c}{Bridge} & & \multicolumn{3}{c}{Aeroplane} \\
		\hline 
		& Deck & Supports & & Wings & Fuselage & Landing Gear \\
		\hline 
		Length (m) & 100 & 15 & & 8 & 20 & 4 \\
		Width (m)  & 15  & 2  & & 3 & 2 & 0.5 \\
		Height (m) & 2   & 2  & & 0.5 & 2 & 0.5 \\
		\hline  
		Young's Modulus (Pa) & $30\times10^9$ & $5\times10^9$ & & $69\times10^9$ & $69\times10^9$ & $69\times10^9$ \\
		Mass Density (kg/$\text{m}^3$) & 2400 & 2000 & & 2700 & 2700 & 2700 \\
		Boundary stiffness (N/m) & \multicolumn{2}{c}{$10\times10^{10}$} & & \multicolumn{3}{c}{$0$}\\
		\hline 
	\end{tabular}
\end{table}

\subsection{Transfer approach}

\begin{figure}[h!]
	\vspace{0.5cm}
	\centering
	\includegraphics[width=0.75\textwidth]{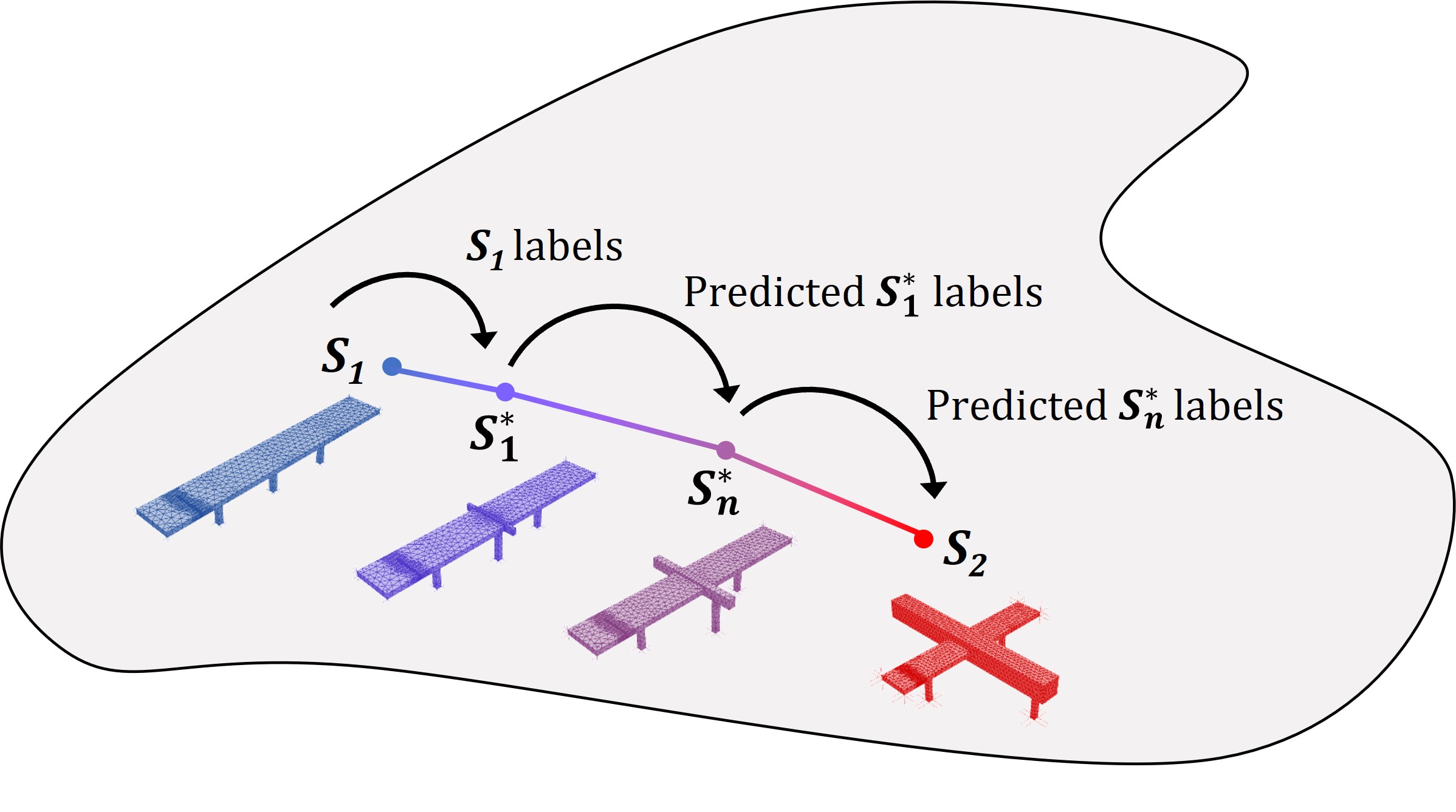}
	\caption{Heterogeneous transfer approach via intermediate structures.}
	\label{fig:intermediate_structures}
	% \vspace{-0.5cm}
\end{figure}

Transfer from the source bridge, $ \boldsymbol{S}_1 $, to the target aeroplane, $ \boldsymbol{S}_2 $, was achieved by incrementally transferring across structures assumed to lie in an intermediate space between them. With the exception of the bridge $ \boldsymbol{S}_1 $, which was assumed to have fully-labelled data, each structure (including intermediates), was assumed to have some labelled normal-condition data and fully unlabelled damage-condition data, where the task was to correctly identify the healthy and damaged datasets. Predicted labels for each structure were then transferred across the chain, where the target for the current transfer became the source for the next, as in Figure \ref{fig:intermediate_structures}. This is a type of self-training, similar to transductive methods in semi-supervised learning \cite{selftraining1967, amini2022}. In the first examples discussed in Section 3.3.1, transfer was performed using normal-condition alignment and an SVM classifier with a linear kernel. In the later examples presented in Section 3.3.2, transfer was performed by first aligning the data, and then using an SVM classifier with a geodesic flow kernel. In each case, 1000 tests were run using randomised starting seeds, to vary both the generated datasets and training data for the classifiers. Transfer accuracy was evaluated by considering the prediction at the end of the chain, and comparing it to that when transferring directly from $ \boldsymbol{S}_1 $ to $ \boldsymbol{S}_2 $. 

%\FloatBarrier
%\FloatBarrier

\subsection{Results}

As previously stated, in the first examples, transfer was performed using normal-condition alignment and an SVM classifier with a linear kernel. In the later examples, transfer was performed by first aligning the data, and then using an SVM classifier with a geodesic flow kernel. In all cases, transfer was performed first using one intermediate structure, and then again using three, 13, and 78 intermediate structures (in other words, using all generated models).

\subsubsection{Linear kernel}

Using normal-condition alignment and an SVM with a linear kernel to transfer via one intermediate structure (shown as $S_n^*$ in Figure \ref{fig:intermediate_structures}), the prediction accuracy for damage labels at the end of the chain was higher than that when transferring directly from $ \boldsymbol{S}_1 $ to $ \boldsymbol{S}_2 $, for 58.9\% of all iterations, and  the average prediction accuracy for damage labels was 23.7\%, compared to 11.3\% for direct transfer. Then, using three intermediate structures (not shown but relatively equally-spaced between the source and target) the prediction accuracy for damage labels at the end of the chain was higher than that when transferring directly from $ \boldsymbol{S}_1 $ to $ \boldsymbol{S}_2 $, for 81.7\% of all iterations. Likewise, the average prediction accuracy for damage labels using three intermediate structures was 48.7\%. Next, using 13 intermediate structures (not shown but relatively equally-spaced between the source and target) the prediction accuracy for damage labels at the end of the chain was higher than that when transferring directly from $ \boldsymbol{S}_1 $ to $ \boldsymbol{S}_2 $, for 96.5\% of all iterations. Likewise, the average prediction accuracy for damage labels using 13 intermediate structures was 88.9\%. Finally, using 78 intermediate structures (as shown in Figure \ref{fig:intermediate_structures}), the prediction accuracy for damage labels at the end of the chain was higher than that when transferring directly from $ \boldsymbol{S}_1 $ to $ \boldsymbol{S}_2 $, for 100.0\% of iterations. The average prediction accuracy for damage labels using 78 intermediate structures was 99.5\%. Using the linear kernel, mean prediction accuracies, either at the end of the chain or directly from $ \boldsymbol{S}_1 $ to $ \boldsymbol{S}_2 $, are shown in Table \ref{tab:Results}. In the table, note that `IS' refers to intermediate structure.

\begin{table}[h]
	\renewcommand{\arraystretch}{1.5}
	\caption{\label{tab:Results}Mean prediction accuracy.}
	\begin{center}
		\begin{tabular}{ p{1.15cm} p{1.15cm} p{1.15cm} p{1.15cm} p{1.15cm} p{0.15cm} p{1.15cm} p{1.15cm} p{1.15cm} p{1.15cm} p{1.15cm} }
			\hline
			\multicolumn{5}{c}{Linear kernel} & &  \multicolumn{5}{c}{Geodesic flow kernel} \\
			\hline
			direct & 1IS & 3IS & 13 IS & 78 IS & &  direct & 1IS & 3IS & 13 IS & 78 IS \\
			\hline
			11.3\% & 23.7\% & 48.7\% & 88.9\% & 99.5\% & & 64.5\% & 88.3\% & 98.8\% & 100.0\% & 100.0\% \\
			\hline
		\end{tabular}
	\end{center}
\end{table}

\subsubsection{Geodesic flow kernel}

Using normal-condition alignment followed by SVM with the geodesic flow kernel, and transferring directly from $ \boldsymbol{S}_1 $ to $ \boldsymbol{S}_2 $, resulted in better transfer 100.0\% of the time, compared to transferring directly from $ \boldsymbol{S}_1 $ to $ \boldsymbol{S}_2 $ using the geodesic flow kernel. Specifically, the average prediction accuracy for damage labels for direct transfer using the geodesic flow kernel was 64.5\%. Good results were achieved using only one intermediate structure (shown as $S_n^*$ in Figure \ref{fig:intermediate_structures}), with a prediction accuracy for damage labels higher than direct transfer 99.9\% of the time relative to direct transfer with the linear kernel and 88.0\% relative to direct transfer with the geodesic flow kernel. The average prediction accuracy for damage labels using one intermediate structure and the geodesic flow kernel was 88.3\%. Good results were also achieved using three intermediate structures (not shown but relatively equally-spaced between the source and target), with a prediction accuracy for damage labels higher than direct transfer 100.0\% of the time relative to direct transfer with the linear kernel and 99.5\% relative to direct transfer with the geodesic flow kernel. The average prediction accuracy for damage labels using three intermediate structures and the geodesic flow kernel was 98.8\%. Excellent results were achieved using 13 intermediate structures (again, these are not shown but relatively equally-spaced between the source and target) the prediction accuracy for damage labels at the end of the chain and the geodesic flow kernel, with a prediction accuracy for damage labels higher than direct transfer for 100.0\% of the iterations, relative to direct transfer with either the linear or geodesic flow kernels. The average prediction accuracy for damage labels using 13 intermediate structures and the geodesic flow kernel was 100.0\%. Using 78 intermediate structures, the transfer prediction accuracy exceeded that from direct transfer for 100.0\% of the iterations, relative to direct transfer with either the linear or geodesic flow kernels. The average prediction accuracy for damage labels using 78 intermediate structures was 100.0\%. Using the geodesic flow kernel, mean prediction accuracy, either at the end of the chain or directly from $ \boldsymbol{S}_1 $ to $ \boldsymbol{S}_2 $, are also shown in Table \ref{tab:Results}.

\section{Conclusions}

This work presents case studies for a novel heterogeneous transfer approach for PBSHM, with formulations based in differential geometry. Using an set of simulated intermediate structures to bridge the gap in information between the source and target, positive transfer was achieved between a simulated `bridge' and `aeroplane', with overall better predictions than direct transfer.

%%%%%%%%%%%%%%%%%%%%
% Acknowledgements %
%%%%%%%%%%%%%%%%%%%%

\section*{Acknowledgements}

TAD, LAB, ND, and KW gratefully acknowledge the support of the UK Engineering and Physical Sciences Research Council (EPSRC), via grant reference EP/W005816/1. LAB is supported by The Mitchell Fellowship in Statistics and Data Analytics from the University of Glasgow. For the purpose of open access, the authors have applied a Creative Commons Attribution (CC BY) licence to any Author Accepted Manuscript version arising.

%%%%%%%%%%%%%%
% References %
%%%%%%%%%%%%%%

% Include your references below, e.g. by appending a '.bib' file. 
% The style, IEEE, is already pre-defined in the class description.

\bibliography{References}

% If BibTeX is not used, you can enter each bibitem separately using the procedure below, which complies with the IEEE reference style. (When you have less than 10 references, use: "\begin{thebibliography}{1}". Switch to "\begin{thebibliography}{10}" is you have 10 or more references.)

	%\begin{thebibliography}{1}
	%	
	%	\bibitem{heylen1997modal}
	%	W. Heylen, S. Lammens and P. Sas,
	%	\textit{Modal Analysis Theory and Testing}.
	%	Leuven, Belgium: Katholieke Universiteit Leuven, Departement Werktuigkunde, 1997.
	%	
	%	\bibitem{sas1995active}
	%	P. Sas, C. Bao, F. Augusztinovicz and W. Desmet,
	%	``Active control of sound transmission through a double panel partition,''
	%	\textit{Journal of Sound and Vibration}, vol. 180, no. 4, pp. 609-625, 1995.
	%	
	%	\bibitem{boonen2001modified}
	%	R. Boonen and P. Sas,
	%	``Modified smith compensation for feedback active noise control in ducts,''
	%	in \textit{Proceedings of the 2001 International Congress and Exhibition on Noise Control Engineering},
	%   The Hague, The Netherlands, 2001, pp. 619-624.
	%\end{thebibliography}

\end{document}